\documentclass{article}

     \PassOptionsToPackage{numbers, compress}{natbib}
     \usepackage[final]{neurips_2020}

\usepackage[utf8]{inputenc} % allow utf-8 input
\usepackage[T1]{fontenc}    % use 8-bit T1 fonts
\usepackage{hyperref}       % hyperlinks
\usepackage{url}            % simple URL typesetting
\usepackage{booktabs}       % professional-quality tables
\usepackage{amsfonts}       % blackboard math symbols
\usepackage{nicefrac}       % compact symbols for 1/2, etc.
\usepackage{microtype}      % microtypography
\usepackage{amsmath}
\usepackage{amsthm}
\usepackage{mathtools}
\usepackage{bbm}

\newtheorem{theorem}{Theorem}
\newtheorem{prop}[theorem]{Proposition}
\newtheorem{lemma}{Lemma}

\newcommand{\EE}[1]{\mathbb{E}\left[#1\right]}
\newcommand\given[1][]{\:#1\vert\:}
\newcommand{\ones}{\mathbbm{1}}

\newcommand{\bx}{\mathbf{x}}
\newcommand{\cX}{\mathcal{X}}
\newcommand{\cA}{\mathcal{A}}
\newcommand{\hp}{\hat{p}}
\newcommand{\tV}{\Tilde{V}}
\newcommand{\bV}{\Bar{V}}

\newif\ifappendix
\appendixtrue

\title{Optimal Policies for the Homogeneous Selective Labels Problem}

\author{%
  Dennis Wei\\
  IBM Research\\
  Yorktown Heights, NY 10598\\
  \texttt{dwei@us.ibm.com} \\
}

\begin{document}

\maketitle

\begin{abstract}
  Selective labels are a common feature of consequential decision-making applications, referring to the lack of observed outcomes under one of the possible decisions. This paper reports work in progress on learning decision policies in the face of selective labels. The setting considered is both a simplified homogeneous one, disregarding individuals' features to facilitate determination of optimal policies, and an online one, to balance costs incurred in learning with future utility. For maximizing discounted total reward, the optimal policy is shown to be a threshold policy, and the problem is one of optimal stopping. In contrast, for undiscounted infinite-horizon average reward, optimal policies have positive acceptance probability in all states. Future work stemming from these results is discussed.
\end{abstract}

\section{Introduction}
\label{sec:intro}

The problem of \emph{selective labels} is common to many consequential decision-making scenarios affecting human subjects. In these scenarios, individuals receive binary decisions, which will be referred to generically as acceptance or rejection. If the decision is to accept, then an outcome label is observed, which determines the utility of the decision. However if the decision is to reject, no outcome is observed. In lending for example, the decision is whether to offer or deny the loan, and the outcome of repayment or default is observed only if the loan is made. In pre-trial bail decisions, the outcome is whether a defendant returns to court without committing another offense, but is not observed if bail is denied. In hiring, a candidate's job performance is observed only if they are hired.

The prevalence and challenges of selective labels were recently highlighted in \cite{lakkaraju2017selective}, which studied the evaluation of machine learning models in comparison to human decision-makers using data labelled selectively by the human decisions themselves. The subject of the present paper is the \emph{learning} of decision policies in the face of selective labels. This problem was addressed indirectly in \cite{dearteaga2018learning}, which proposed label imputation in regions where humans are highly confident, and more directly and deeply by \citet{kilbertus2020fair}. In \cite{kilbertus2020fair}, the goal is to maximize expected utility (possibly including a fairness penalty) over a held out population, given data and labels collected selectively by a suboptimal existing policy. \citeauthor{kilbertus2020fair} showed that an existing policy that is deterministic, commonly achieved by thresholding the output of a predictive model, may condemn future policies to suboptimality. However, if the existing policy is stochastic and ``exploring'', then the optimal policy can be learned and a stochastic gradient ascent algorithm is proposed to do so.

This paper reports work in progress that takes a step back from the setting of \cite{kilbertus2020fair}, considering a simpler case in which individuals are assumed to be drawn from a homogeneous population, without features to distinguish them. At the same time, an online version of the problem is formulated that accounts for the costs of decisions taken during learning, unlike in \cite{kilbertus2020fair} where these costs do not enter into the objective. 
The focus on the simpler homogeneous setting attempts to get at the essence of the problem, and its benefit is that the structure of the optimal acceptance policy can be determined. This is done by formulating the problem as a partially observable Markov decision process (POMDP) and applying dynamic programming.

In the case of discounted total reward in Section~\ref{sec:homo:disc}, the optimal policy is shown to be a threshold policy, and the problem moreover is one of \emph{optimal stopping}. Properties of the optimal value functions are derived, showing that the policy becomes more stringent (i.e., the stopping/rejection set grows) as more observations are collected. Furthermore, the dynamic programming recursion provides an efficient way to approximate the optimal policy computationally. Section~\ref{sec:discuss} discusses the potential utility of these findings for the more general selective labels problem with features.

Section~\ref{sec:homo:avg} briefly considers the case of undiscounted average reward over an infinite horizon. Here it is found that the policy should accept individuals with positive probability regardless of the belief state. This is in line with the exploring policies in \cite{kilbertus2020fair} and contrasts sharply with the case of discounted total reward in Section~\ref{sec:homo:disc}.

\paragraph{Other related work}
The selective labels problem is related to policy learning \cite{dudik2011doubly,swaminathan2015batch,athey2017policy,kallus2018balanced}, 
causal inference \cite{hernan2020causal}, and multi-arm (contextual) bandits \cite{agarwal2014taming,joseph2016fairness}, in that only the outcome resulting from the selected action is observed. It is distinguished by there being no observation at all in the case of rejection. 
Notwithstanding this difference, it appears possible to view the online formulation considered herein as a simpler special kind of bandit problem, as noted in Section~\ref{sec:homo:disc}. This simplicity makes it amenable to an optimal dynamic programming approach as opted for in this paper.

Selective labels and similar limited feedback phenomena
have been considered in the literature on machine learning for consequential decision-making. \citet{kallus2018residual} study similar censoring of data by an existing policy, the ``residual unfairness'' of supposedly fair policies learned from this data, and corrected measures of fairness. The lending scenario mentioned in the introduction is the running example used in \cite{liu2018delayed} (\cite{mouzannar2019from} is similar), and the structural causal models of \cite{creager2020causal} make clear that the loan outcome is really a \emph{potential} outcome. In predictive policing \cite{lum2016predict}, crimes are discovered by police only in areas where they are deployed, which can lead to runaway feedback loops but is also correctable by importance sampling \cite{ensign2018runaway}.

\section{General problem formulation}
\label{sec:general}

The general problem of selective labels that is the eventual goal of this paper is as follows: Individuals $i = 0, 1, \dots$ arrive sequentially with features $\bx_i \in \cX$. They also have sensitive attributes indicating group membership $a_i \in \cA$, if we are to consider fairness with respect to these groups. A decision of \emph{accept} ($t_i = 1$) or \emph{reject} ($t_i = 0$) is made based on each individual's $\bx_i$ and $a_i$ according to a \emph{decision policy} $\Pi: \cX \times \cA \mapsto [0,1]$, where $\Pi(\bx,a) = \Pr(t=1 \given \bx,a)$ is the probability of acceptance. The policy is thus permitted to be stochastic, although it will be seen that this is not needed in the case of discounted total reward. If the decision is accept, then a binary outcome $y_i$ is observed, with $y_i = 1$ representing \emph{success} and $y_i = 0$ \emph{failure}. If the decision is reject, then no outcome is observed, hence the term \emph{selective labels}. Individuals' features, sensitive attributes, and outcomes are independently and identically distributed according to a joint distribution $p(\bx, a, y) = p(y\given \bx, a) p(\bx, a)$.

Decisions and outcomes incur rewards according to $t_i (y_i - c)$ for $c \in (0,1)$, following the formulation of \cite{kilbertus2020fair,corbett-davies2017algorithmic}, i.e., a reward of $1-c$ if acceptance leads to success, $-c$ if acceptance leads to failure, and $0$ if the individual is rejected.  
As noted in \cite{kilbertus2020fair}, the cost of rejection, say $(1-t_i) g(y_i)$, is \emph{unknowable} because of unobserved outcomes (although it is surmised to be negative for the individual) except possibly for constant $g$, which can then be set to zero 
without loss of generality. The parameter $c$ represents the relative reward/cost of success/failure and can reflect those of both the individual as well as the decision-maker. For example in the lending scenario, the decision-maker's (lender's) rewards are fairly clear: interest earned in the case of success (repayment), and loss of principal (or some expected fraction thereof) in the case of failure (default). Individual rewards may also be taken into account although harder to quantify, for example the value of accomplishing the objective of the loan (e.g.~owning a home) in the case of success, or damage to creditworthiness in the case of failure \cite{liu2018delayed}, over and above the loss $g$ due to denial of the loan.

The objective of \emph{utility} is quantified by the expectation of the discounted infinite sum of rewards, 
\begin{equation}\label{eqn:utilDisc}
    \EE{ \sum_{i=0}^{\infty} \gamma^i t_i (y_i - c)} = \EE{ \sum_{i=0}^{\infty} \gamma^i \pi(\bx_i,a_i) (p(y_i=1\given \bx_i,a_i) - c)}
\end{equation}
for some discount factor $\gamma < 1$. The right-hand side results from taking the conditional expectation given $(\bx_i, a_i)$, leaving an expectation over $(\bx_i,a_i) \sim p(\bx,a)$. The undiscounted average reward over an infinite horizon, 
\begin{equation}\label{eqn:utilAvg}
    \lim_{N\to\infty} \EE{ \frac{1}{N} \sum_{i=0}^{N-1} \pi(\bx_i,a_i) (p(y_i=1\given \bx_i,a_i) - c)},
\end{equation}
will also be considered in Section~\ref{sec:homo:avg}. A fairness objective can also be formulated as in \cite{kilbertus2020fair} but this will not be considered herein. 

The expectation with respect to $(\bx, a)$ on the right-hand side of \eqref{eqn:utilDisc} and in \eqref{eqn:utilAvg} indicates that the problem of determining policy $\Pi(\bx, a)$ can be decomposed (at least conceptually) over values of $(\bx,a)$. This is clearest in the case of discrete domains $\cX$ and $\cA$ for which the expectation is a sum, weighted by $p(\bx,a)$. 
The decomposition 
motivates in part the study of a simpler problem in which $(\bx, a)$ is dropped (or fixed), resulting in a homogeneous population. This simplified ``homogeneous'' problem is the subject of the remainder of the paper.

\section{The homogeneous problem}
\label{sec:homo}

To simplify notation for the homogeneous problem, define success probability $p \coloneqq p(y_i = 1)$ and acceptance probability $\pi$ given by policy $\Pi$ (the inputs to $\Pi$ are left unspecified for the moment). Then the expected immediate reward per individual is $\pi (p - c)$. Section~\ref{sec:homo:disc} addresses the case of discounted total reward, $\sum_{i=0}^\infty \gamma^i \pi (p - c)$ analogous to \eqref{eqn:utilDisc}, while Section~\ref{sec:homo:avg} addresses undiscounted average reward \eqref{eqn:utilAvg}.  

\subsection{Discounted total reward}
\label{sec:homo:disc}

If the success probability $p$ is known, then the solution that maximizes all of the reward functions is immediate: $\Pi^*(p) = \ones(p > c)$, where $\ones(\cdot)$ is the indicator function that yields $1$ when its argument is true. The optimal discounted total reward is therefore the following function of $p$:
\begin{equation}\label{eqn:V*inf}
    V^*(\infty, p) = \sum_{i=0}^{\infty} \gamma^i \max\{p-c, 0\} = \frac{1}{1-\gamma} \max\{p-c, 0\}.
\end{equation}
As will be explained more fully later, the $\infty$ in $V^*(\infty, p)$ denotes exact knowledge of $p$, i.e.~from an infinite sample.

The challenge of course is that $p$ is not known but must be learned as decisions are made. The approach taken herein is to regard the case of known $p$ as a Markov decision process (MDP) with state $p$ and no dynamics (i.e.~$p_{i+1} = p_i$). The case of unknown $p$ is then treated as the corresponding partially observable MDP (POMDP) using a \emph{belief state} for $p$ \cite[Sec.~5.4]{bertsekas2005dynamic}. An alternative approach may be to treat the homogeneous problem as a special kind of two-arm bandit problem (and the general formulation in Section~\ref{sec:general} as the corresponding contextual bandit), where the rewards of the reject arm are unknowable and thus taken to be zero. Here it is shown that the POMDP approach allows the optimal policy to be obtained through dynamic programming.

To define the belief state, a beta distribution prior is placed on $p$: $p \sim B(s_0, n_0-s_0)$, where the shape parameters $\alpha = s_0$, $\beta = n_0 - s_0$ are expressed for convenience in terms of a number $s_0$ of ``virtual successes'' in $n_0$ virtual observations. Since $p$ is the parameter of a Bernoulli random variable, the beta distribution is a conjugate prior. It follows that the posterior distribution of $p$ before individual $i$ arrives, given $n'_i = \sum_{j=0}^{i-1} t_j$ outcomes and $s'_i = \sum_{j=0}^{i-1} t_j y_j$ successes observed thus far, is also beta with parameters $\alpha = s_0 + s'_i$ and $\beta = (n_0 + n'_i) - (s_0 + s'_i)$. 
Thus we define the pair $(n_i, s_i)$ as the belief state for $p$,
\begin{equation}\label{eqn:BSrecursion}
    n_i = n_0 + \sum_{j=0}^{i-1} t_j = n_{i-1} + t_{i-1}, \qquad s_i = s_0 + \sum_{j=0}^{i-1} t_j y_j = s_{i-1} + t_{i-1} y_{i-1},
\end{equation}
and make the acceptance policy a function thereof, $\Pi(n_i, s_i)$. 
From the recursive definition in \eqref{eqn:BSrecursion}, given state $(n_i, s_i)$ and action (acceptance probability) $\pi_i = \Pi(n_i,s_i)$, the next state is given by 
\begin{equation}\label{eqn:BSdynamics}
(n_{i+1}, s_{i+1}) = \begin{cases}
(n_i+1, s_i+1) & \text{with probability } \pi_i \hp_i \text{ and reward } 1-c,\\
(n_i+1, s_i) & \text{with probability } \pi_i (1 - \hp_i) \text{ and reward } -c,\\
(n_i, s_i) & \text{with probability } 1 - \pi_i \text{ and reward } 0,
\end{cases}
\end{equation}
where $\hp_i \coloneqq s_i / n_i = \EE{p\given n_i,s_i}$ is the success probability marginalized over the posterior. The three cases in \eqref{eqn:BSdynamics} correspond to acceptance and success, acceptance and failure, and rejection.

The initial state $(n_0, s_0)$, i.e.~the parameters of the beta prior, can be chosen based on initial beliefs about $p$. This choice is clearer when outcome data has already been collected by an existing policy. In this case, $n_0$ can be the number of outcomes observed, and $s_0$ the number of successes.

We now derive the dynamic programming recursion that specifies the optimal policy. Denote by $V^{\Pi}(n, s)$ the value function at state $(n,s)$ under policy $\Pi$, i.e., the expected discounted sum of rewards from following policy $\Pi$. The sample index $i$ is dropped henceforth because the dependence is on $(n,s)$, irrespective of the number of samples used to attain this state. In particular, we write $\pi = \Pi(n,s)$. From the state transitions and rewards in \eqref{eqn:BSdynamics}, we have
\[
V^\Pi(n,s) = \pi (\hp - c) + \gamma \left[ \pi \hp V^\Pi(n+1,s+1) + \pi (1-\hp) V^\Pi(n+1,s) + (1-\pi) V^\Pi(n,s) \right],
\]
where the first term is the immediate reward and the quantity in square brackets is the expected future reward, discounted by $\gamma$. Solving the previous equation for $V^\Pi(n,s)$ yields 
\[
    V^\Pi(n,s) = \frac{\pi}{1 - \gamma + \gamma\pi} \left(\hp - c + \gamma \left[\hp V^\Pi(n+1,s+1) + (1-\hp) V^\Pi(n+1,s) \right]\right).
\]
It will be more convenient to re-parametrize the state in terms of $n$ and $\hp = s/n$. Thus we have 
\begin{equation}\label{eqn:Vpi}
V^\Pi(n,\hp) = \frac{\pi}{1 - \gamma + \gamma\pi} \left(\hp - c + \gamma \left[ \hp V^\Pi\left(n+1, \frac{n\hp + 1}{n+1}\right) + (1-\hp) V^\Pi\left(n+1, \frac{n\hp}{n+1}\right) \right]\right).
\end{equation}

An optimal policy is obtained recursively by assuming that it is followed from state $n+1$ onward, which replaces $V^\Pi(n+1,\cdot)$ by the optimal value $V^*(n+1,\cdot)$, and then maximizing the right-hand side of \eqref{eqn:Vpi} with respect to the current action $\pi$ \cite{bertsekas2005dynamic}:
\begin{equation}\label{eqn:V*1}
    V^*(n,\hp) = \max_{\pi \in [0,1]} \frac{\pi}{1 - \gamma + \gamma\pi} \underbrace{\left(\hp - c + \gamma \left[ \hp V^*\left(n+1, \frac{n\hp + 1}{n+1}\right) + (1-\hp) V^*\left(n+1, \frac{n\hp}{n+1}\right) \right]\right)}_{\tV(n,\hp)}.
\end{equation}
The key observation is that the dependence on $\pi$ is confined to the first factor above and is moreover monotonically increasing or decreasing depending on the sign of $\tV(n,\hp)$. This leads to the following conclusion.
\begin{theorem}\label{thm:disc}
The optimal acceptance policy that maximizes discounted total reward is a threshold policy: $\Pi^*(n,\hp) = \ones(\tV(n,\hp) > 0)$, where $\tV(n,\hp)$ is given in \eqref{eqn:V*1}.
\end{theorem}
It follows that $V^*(n,\hp) = \max\{\tV(n,\hp), 0\}$, i.e., $V^*(n,\hp)$ satisfies the recursion 
\begin{equation}\label{eqn:V*2}
    V^*(n,\hp) = \max\left\{ \hp - c + \gamma \left[\hp V^*\left(n+1, \frac{n\hp + 1}{n+1}\right) + (1-\hp) V^*\left(n+1, \frac{n\hp}{n+1}\right) \right], 0 \right\}.
\end{equation}

Theorem~\ref{thm:disc} shows that the optimal policy for discounted total reward does not require stochasticity, as claimed earlier. It also shows that the problem is one of \emph{optimal stopping} \cite[Sec.~4.4]{bertsekas2005dynamic}: in each state $(n,\hp)$, there is the option ($\pi = 0$) to stop accepting and thus stop observing, which freezes the state at $(n,\hp)$ thereafter with zero reward. The decision to stop is based on the sign of $\tV(n,\hp)$, the optimal expected reward from continuing. The optimal policy is thus characterized by the \emph{stopping set}, the set of $(n,\hp)$ at which it is optimal to stop, and continue otherwise.

In the limiting case as $n \to \infty$, $V^*(n,\hp)$ and $\Pi^*(n,\hp)$ are known explicitly. This is because $\hp$ converges to the true success probability $p$, by the law of large numbers. We therefore have $\Pi^*(\infty, \hp) = \ones(\hp > c)$ and $V^*(\infty, \hp)$ as given in \eqref{eqn:V*inf}, explaining the previous notation. The corresponding stopping set is the interval $[0, c]$.

For finite $n$, a natural way of approximating $V^*(n,\hp)$ and $\Pi^*(n,\hp)$ is as follows: Choose a large integer $N$, which will also index the approximation, $V^N(n,\hp)$, and set $V^N(N+1, \hp) = V^*(\infty, \hp)$, the infinite-sample value function \eqref{eqn:V*inf}. Then use \eqref{eqn:V*2} with $V^{N}$ in place of $V^*$ to recursively compute $V^{N}(n,\hp)$ for $n = N, N-1, \dots$ and $\hp = 0, 1/n, 2/n, \dots, 1$. The corresponding policy is $\Pi^{N}(n,\hp) = \ones(V^{N}(n,\hp) > 0)$.

\begin{figure}[ht]
  \centering
  \includegraphics[width=0.48\textwidth]{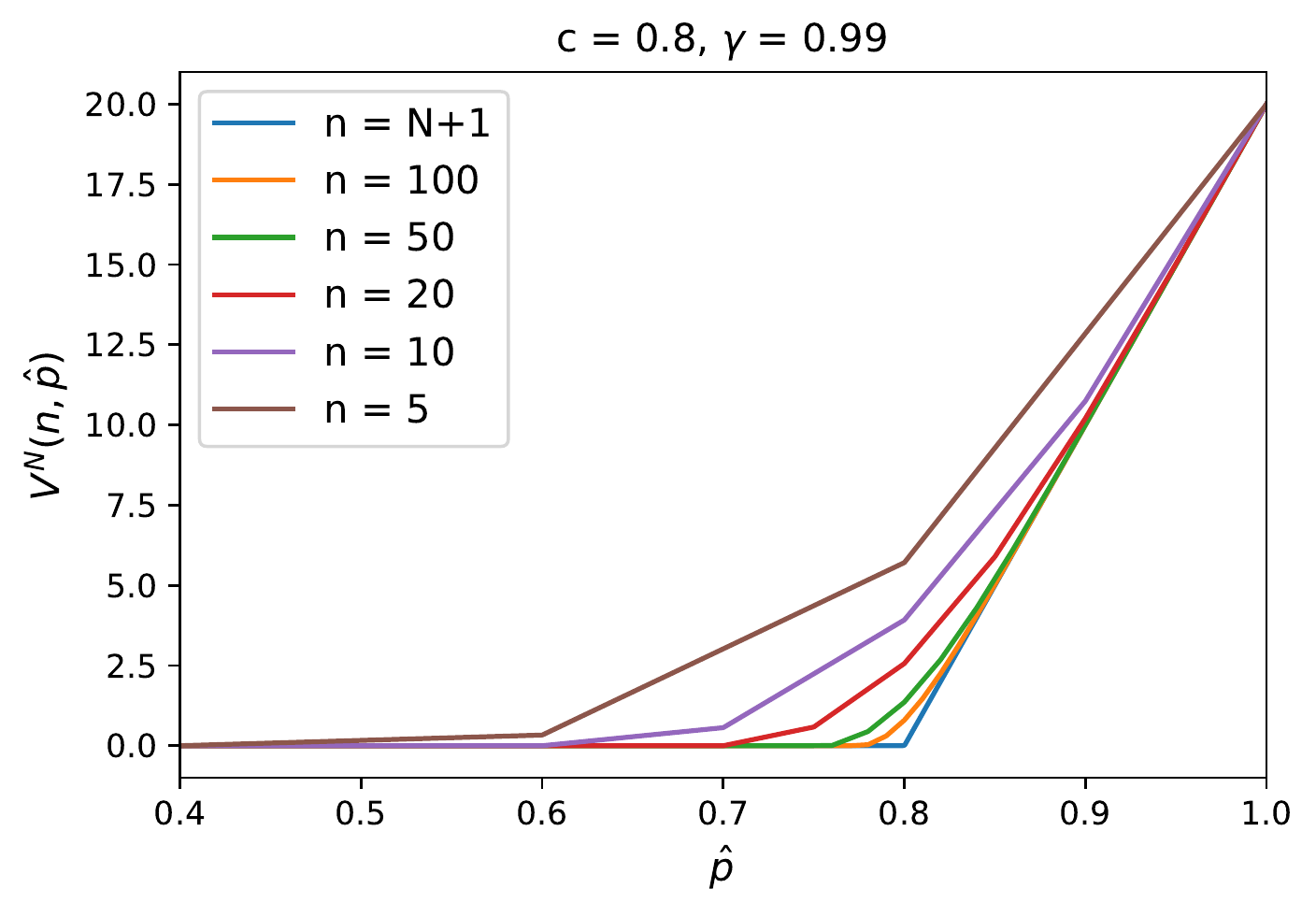}
  \includegraphics[width=0.48\textwidth]{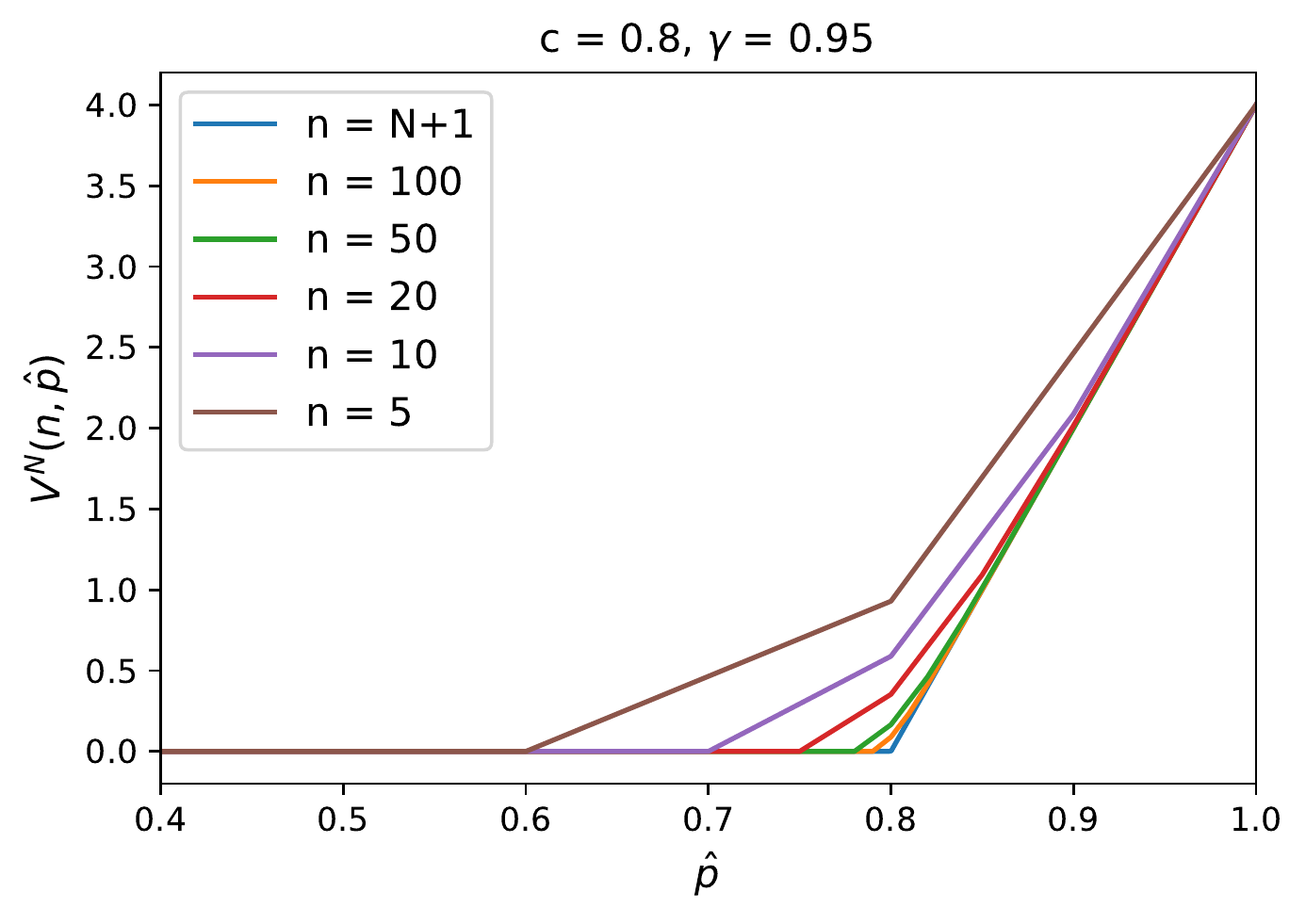}
  \caption{Optimal value function approximations $V^N(n,\hp)$ for $N = 1000$.}
  \label{fig:Vstar}
\end{figure}

Figure~\ref{fig:Vstar} plots the result of the above computation for $N = 1000$, $c = 0.8$, and $\gamma \in \{0.99,0.95\}$. As noted above, $V^N(n,\hp)$ is computed only for $\hp$ equal to integer multiples of $1/n$ and is then linearly interpolated for visualization. With this caveat in mind, the plots do suggest that $V^N(n,\hp) \geq V^N(n+1,\hp)$ and that $V^N(n,\hp)$ is a non-decreasing convex function of $\hp$ for all $n$. They also show that $V^N(100,\hp)$ is close to $V^N(1001,\hp) = V^*(\infty,\hp)$ and that the differences become progressively larger as $n$ decreases. This suggests that $N$ does not need to be very large to approximate $V^*(n,\hp)$ well. It is left to future work however to make this precise by analyzing the approximation error.
 
The monotonicity and convexity properties suggested by Figure~\ref{fig:Vstar} do in fact hold generally. 
\begin{prop}\label{prop:V*p}
The optimal value function $V^*(n,\hp)$ is non-decreasing and convex in $\hp$ for all $n$.
\end{prop}
\begin{prop}\label{prop:V*n}
The optimal value function $V^*(n,\hp)$ is non-increasing in $n$, i.e.~$V^*(n,\hp) \geq V^*(n+1,\hp)$ for all $\hp \in [0,1]$.
\end{prop}
Monotonicity in both $\hp$ and $n$ implies that the stopping set in state $n$, $\{\hp: V^*(n,\hp) \leq 0\}$, is an interval $[0, c_n]$ that shrinks as $n$ decreases, $c_n \leq c_{n+1} \leq \dots \leq c$. In other words, the acceptance policy is more lenient in early stages and gradually approaches the policy when $p$ is known.

The proof structure for Propositions~\ref{prop:V*p} and \ref{prop:V*n} is described below, deferring 
the algebra to Appendix~%
\ifappendix
    \ref{sec:proofs}.
\else
    A.
\fi
Both are proven by induction over decreasing $n$. Technically, the proofs are only for the approximations $V^{N}(n,\hp)$ to $V^*(n,\hp)$ described above. However by taking $N \to \infty$, $V^N \to V^*$ and the properties extend to $V^*$ as well. The proofs also require the following lemma proven in Appendix~%
\ifappendix
    \ref{sec:proofs:chords}.
\else
    A.1.
\fi
\begin{lemma}\label{lem:chords}
Let $f: \mathbb{R} \mapsto \mathbb{R}$ be convex. Then for any $x$ and $\alpha \in [0,1]$, $\alpha f(x + (1-\alpha)\delta) + (1-\alpha) f(x - \alpha\delta)$ is non-decreasing in $\delta \geq 0$.
\end{lemma}
%The algebraic proof of Lemma~\ref{lem:chords} is in Appendix~\ref{sec:proofs:chords} but it can be proved graphically by drawing chords that intersect $f(x)$ at $x - \alpha\delta$ and $x + (1-\alpha)\delta$ for increasing $\delta$.

For Proposition~\ref{prop:V*p}, the base case is $n = N+1$, for which $V^N(N+1,\hp) = V^*(\infty,\hp) = \max\{\hp-c, 0\} / (1-\gamma)$ is both non-decreasing and convex in $\hp$. Appendix~%
\ifappendix
    \ref{sec:proofs:V*p} 
\else
    A.2 
\fi
proves the inductive step, i.e.~that $V^N(n+1,\hp)$ being non-decreasing and convex in $\hp$ implies the same for $V^N(n,\hp)$, with the help of Lemma~\ref{lem:chords}.

For Proposition~\ref{prop:V*n}, 
the base case requires showing that $V^N(N,\hp) \geq V^N(N+1, \hp) = V^*(\infty,\hp)$, where $V^N(N,\hp)$ is obtained from \eqref{eqn:V*2} with $V^*(\infty,\hp)$ in place of $V^*(n+1, \hp)$. This calculation is shown in Appendix~%
\ifappendix
    \ref{sec:proofs:V*n}.
\else
    A.3.
\fi
The inductive step follows in Appendix~%
\ifappendix
    \ref{sec:proofs:V*n}, 
\else
    A.3, 
\fi
again using Lemma~\ref{lem:chords}.

\subsection{Undiscounted average reward}
\label{sec:homo:avg}

This subsection briefly considers the case of undiscounted average reward \eqref{eqn:utilAvg} in the homogeneous setting. The same POMDP approach is followed to construct a belief state $(n_i, s_i)$ for the unknown success probability $p$. Let $\pi_i = \Pi(n_i, s_i)$ be the acceptance probability given by a policy $\Pi$ for state $(n_i, s_i)$; the expected immediate reward is then $\pi_i (\hp_i - c)$, recalling that $\hp_i = s_i / n_i$. Define 
\begin{equation}\label{eqn:VtoGoAvg}
    V_i^\Pi(n_i, s_i) = \frac{1}{N} \sum_{j=i}^{N-1} \EE{\pi_j (\hp_j - c) \given n_i,s_i}
\end{equation}
to be the sum of rewards, divided by $N$, starting from individual $i$ under policy $\Pi$. We wish to maximize $V_0^\Pi(n_0, s_0)$ in the limit $N \to \infty$. As the number of observations $n_i \to \infty$, we again have $\hp_i \to p$ by the law of large numbers and optimal reward $V_i^*(\infty, \hp) = \max\{\hp-c, 0\}$. 

Equation~\eqref{eqn:VtoGoAvg} can be rewritten as a recursion using the same state transition probabilities as in \eqref{eqn:BSdynamics}:
\[
V_i^\Pi(n_i, s_i) = \frac{1}{N} \pi_i (\hp_i - c) + \pi_i \hp_i V_{i+1}^\Pi(n_i+1, s_i+1) + \pi_i (1-\hp_i) V_{i+1}^\Pi(n_i+1, s_i) + (1-\pi_i) V_{i+1}^\Pi(n_i, s_i).
\]
Taking the limit $N \to \infty$, the first term vanishes and the sample index $i$ again ceases to matter, i.e., $V_{i+1}^\Pi \to V_i^\Pi = V^\Pi$ and the subscript $i$ is dropped elsewhere. The result can be rearranged to yield 
\[
\pi \left(V^\Pi(n,s) - \hp V^\Pi(n+1, s+1) - (1-\hp) V^\Pi(n+1, s) \right) = 0.
\]
There are two cases corresponding to choices of actions: Either $\pi = \Pi(n,s) = 0$, which stops the state evolution and results in zero reward, $V^\Pi(n,s) = 0$, or $\pi > 0$ and the value function satisfies
\begin{equation}\label{eqn:VpiAvg}
    V^\Pi(n,\hp) = \hp V^\Pi\left(n+1, \frac{n\hp + 1}{n+1}\right) + (1-\hp) V^\Pi\left(n+1, \frac{n\hp}{n+1}\right),
\end{equation}
again re-parametrizing the state in terms of $n$ and $\hp$. In Appendix~%
\ifappendix
    \ref{sec:proofs:avg}, 
\else
    A.4,
\fi
it is shown that the choice $\pi > 0$ leads to non-negative value $V^\Pi(n,\hp) \geq 0$ and is hence preferred in all states $(n, \hp)$.

\begin{theorem}\label{thm:avg}
Policies that maximize undiscounted infinite-horizon average reward accept individuals with positive probability $\Pi(n,\hp) > 0$ in all belief states $(n,\hp)$. 
\end{theorem}
Theorem~\ref{thm:avg} shares a similar spirit with the exploring policies in \cite{kilbertus2020fair}, which assign positive acceptance probability to all subsets of $\cX \times \cA$ with positive probability under $p(\bx,a)$. 
It clearly contrasts with Theorem~\ref{thm:disc} for the case of discounted total reward, where stopping sets are optimal.

Theorem~\ref{thm:avg} however does not provide further guidance on selecting a policy. It does not even distinguish between an always-accept policy $\Pi(n,\hp) \equiv 1$ and a stochastic one, $\Pi(n,\hp) = \pi \in (0,1)$, which spends a geometrically distributed amount of time in state $n$ before eventually moving to $n+1$. Intuitively, this seems to be because the lack of a discount factor means that any short-term cost incurred in learning the parameter $p$ is trumped by eventual long-term reward. Indeed, one can conceive of the following two-phase $N$-step policy (with $N \to \infty$): The first ``explore'' phase learns $p$ using a number of samples $N_1$ that increases to infinity but sublinearly in $N$, for example using the always-accept policy $\Pi(n,\hp) \equiv 1$. The second $(N-N_1)$-step phase simply ``exploits'' this knowledge using the threshold policy $\ones(\hp > c)$. Future work could consider the analysis of these and similar policies.

\section{Discussion}
\label{sec:discuss}

Section~\ref{sec:homo:disc} presented the optimal acceptance policy that maximizes discounted total reward for a homogeneous selective labels problem that does not consider features of individuals. A recursive algorithm was also proposed to approximate the optimal policy. While more work remains to analyze the error in this approximation, the algorithm's computational ease is appealing (the full recursion starting from $N=1000$ takes less than a second on a MacBook Pro), and its basis in dynamic programming avoids the need for stochastic exploration to discover the optimal policy. Not only may stochastic exploration take longer to converge, it may also be objectionable for making consequential decisions non-deterministically, as noted in \cite{kilbertus2020fair}. On the other hand, Propositions~\ref{prop:V*p} and \ref{prop:V*n} suggest their own kind of ``sequence unfairness'': early-arriving individuals are subject to a more lenient acceptance policy, enjoying the ``benefit of the doubt'' in the population's true success probability.

It is envisioned that the results in Section~\ref{sec:homo:disc} can be leveraged to solve the more general selective labels problem with features in Section~\ref{sec:general}. Again, this is most apparent if the feature spaces $\cX$ and $\cA$ are discrete and have relatively small cardinality. In this case, the conversion of the expectation in \eqref{eqn:utilDisc} into a sum implies that the optimal solution is to run multiple optimal homogeneous policies in parallel, one for each value of $(\bx, a)$. Indeed, the closest next step may be to consider discrete group membership $A \in \cA$ but no other features. This setting would allow the optimal homogeneous policy to be carried over directly and some group fairness issues to be studied.

If $\cX$ and $\cA$ are not discrete or have cardinalities that are too large, then one approach that seems worth exploring is to combine the optimal homogeneous policy $\Pi^*(n,\hp)$ with modelling of how the quantities $n$ and $\hp$ (or quantities that play a similar role) vary as functions of $(\bx, a)$. In the case of $\hp$, this is the standard probabilistic classification problem of approximating the conditional probability $p(y\given \bx, a)$. The case of $n$ is less clear. Intuitively however, the sample size $n$ represents a kind of confidence in the estimated probability $\hp$, which suggests using a model of predictor confidence.

\begin{ack}
The author thanks Eric Mibuari and Andrea Simonetto for helpful discussions.
\end{ack}

%\small
\bibliography{refs}
\bibliographystyle{plainnat}

\ifappendix
    \clearpage
\appendix

\section{Proofs}
\label{sec:proofs}

\subsection{Proof of Lemma~\ref{lem:chords}}
\label{sec:proofs:chords}

Let $0 \leq \delta_1 \leq \delta_2$. By the convexity of $f$,
\begin{align*}
    f(x + (1-\alpha) \delta_1) &\leq \left(1 - \frac{\delta_1}{\delta_2}\right) f(x) + \frac{\delta_1}{\delta_2} f(x + (1-\alpha) \delta_2),\\
    f(x - \alpha \delta_1) &\leq \left(1 - \frac{\delta_1}{\delta_2}\right) f(x) + \frac{\delta_1}{\delta_2} f(x - \alpha \delta_2).
\end{align*}
Multiplying the first inequality by $\alpha$, the second inequality by $1-\alpha$, and summing, 
\begin{multline*}
    \alpha f(x + (1-\alpha) \delta_1) + (1-\alpha) f(x - \alpha \delta_1)\\
    \leq \left(1 - \frac{\delta_1}{\delta_2}\right) f(x) + \frac{\delta_1}{\delta_2} \left[ \alpha f(x + (1-\alpha) \delta_2) + (1-\alpha) f(x - \alpha \delta_2) \right].
\end{multline*}
Since we also have 
\[
f(x) \leq \alpha f(x + (1-\alpha) \delta_2) + (1-\alpha) f(x - \alpha \delta_2),
\]
the result follows, i.e.
\[
\alpha f(x + (1-\alpha) \delta_1) + (1-\alpha) f(x - \alpha \delta_1) \leq \alpha f(x + (1-\alpha) \delta_2) + (1-\alpha) f(x - \alpha \delta_2).
\]

\subsection{Proof of Proposition~\ref{prop:V*p}: Inductive step}
\label{sec:proofs:V*p}

Here the inductive step is proven, i.e., $V^N(n+1,\hp)$ being non-decreasing and convex in $\hp$ implies that $V^N(n,\hp)$ is also non-decreasing and convex. Since $V^N(n,\hp) = \max\{\tV(n,\hp), 0\}$ and $\tV(n,\hp) = \hp - c + \gamma \bV(n,\hp)$, where 
\[
\bV(n,\hp) = \hp V^N\left(n+1, \frac{n\hp + 1}{n+1}\right) + (1-\hp) V^N\left(n+1, \frac{n\hp}{n+1}\right),
\]
it suffices to show that $\bV(n,\hp)$ is non-decreasing and convex. This is because these properties are preserved under addition with the function $\hp - c$, which is increasing and convex, and under the pointwise maximum with the zero function (also convex).

To show that $\bV(n,\hp)$ is non-decreasing in $\hp$, let $p_1 \leq p_2$. Then
\begin{align*}
    \bV(n,p_2) - \bV(n,p_1) &= p_2 V^N\left(n+1, \frac{np_2 + 1}{n+1}\right) + (1-p_2) V^N\left(n+1, \frac{np_2}{n+1}\right)\\
    &\quad {} - p_1 V^N\left(n+1, \frac{np_1 + 1}{n+1}\right) - (1-p_1) V^N\left(n+1, \frac{np_1}{n+1}\right)\\
    &= (p_2 - p_1) \left[ V^N\left(n+1, \frac{np_2 + 1}{n+1}\right) - V^N\left(n+1, \frac{np_2}{n+1}\right) \right]\\
    &\quad {} + p_1 \left[ V^N\left(n+1, \frac{np_2 + 1}{n+1}\right) + V^N\left(n+1, \frac{np_1}{n+1}\right) \right.\\
    &\quad\qquad \left.{} - V^N\left(n+1, \frac{np_2}{n+1}\right) - V^N\left(n+1, \frac{np_1 + 1}{n+1}\right) \right]\\
    &\quad {} + \left[ V^N\left(n+1, \frac{np_2}{n+1}\right) - V^N\left(n+1, \frac{np_1}{n+1}\right) \right].
\end{align*}
In the final right-hand side above, the first and third quantities in square brackets are non-negative because of the inductive assumption that $V^N(n+1,\hp)$ is non-decreasing in $\hp$. The second bracketed quantity is also shown to be non-negative by applying Lemma~\ref{lem:chords} to $V^N(n+1,\hp)$, assumed to be convex in $\hp$, with 
\[
x = \frac{n(p_2 + p_1) + 1}{2(n+1)}, \quad \alpha = \frac{1}{2}, \quad \delta_1 = \frac{n(p_2 - p_1) - 1}{2(n+1)} \leq \delta_2 = \frac{n(p_2 - p_1) + 1}{2(n+1)}.
\]
Thus $\bV(n,p_2) - \bV(n,p_1) \geq 0$ as required.

To show that $\bV(n,\hp)$ is convex in $\hp$, we require 
\begin{equation}\label{eqn:VbarConvex1}
    \alpha \bV(n,p_1) + (1-\alpha) \bV(n,p_2) \geq \bV(n, \alpha p_1 + (1-\alpha) p_2)
\end{equation}
for $\alpha \in [0,1]$. The left-hand side yields 
\begin{align}
    &\alpha \bV(n,p_1) + (1-\alpha) \bV(n,p_2)\nonumber\\ 
    &\qquad = \alpha p_1 V^N\left(n+1, \frac{np_1 + 1}{n+1}\right) + (1-\alpha) p_2 V^N\left(n+1, \frac{np_2 + 1}{n+1}\right)\nonumber\\
    &\qquad\quad {} + \alpha (1-p_1) V^N\left(n+1, \frac{np_1}{n+1}\right) + (1-\alpha) (1-p_2) V^N\left(n+1, \frac{np_2}{n+1}\right)\nonumber\\
    &\qquad \geq (\alpha p_1 + (1-\alpha) p_2) V^N\left(n+1, \frac{n}{n+1} \frac{\alpha p_1^2 + (1-\alpha) p_2^2}{\alpha p_1 + (1-\alpha) p_2} + \frac{1}{n+1} \right)\nonumber\\
    &\qquad\quad {} + (1 - \alpha p_1 + (1-\alpha) p_2) V^N\left(n+1,  \frac{n}{n+1} \frac{\alpha (1-p_1) p_1 + (1-\alpha) (1-p_2) p_2}{1 - \alpha p_1 + (1-\alpha) p_2} \right)\label{eqn:VbarConvex2}
\end{align}
where the convexity of $V^N(n+1,\hp)$ has been applied separately to the second line and third line above (note $\alpha (1-p_1) + (1-\alpha) (1-p_2) = 1 - \alpha p_1 + (1-\alpha) p_2$). The right-hand side of \eqref{eqn:VbarConvex1} is 
\begin{multline}\label{eqn:VbarConvex3}
\bV(n, \alpha p_1 + (1-\alpha) p_2) = (\alpha p_1 + (1-\alpha) p_2) V^N\left(n+1, \frac{n (\alpha p_1 + (1-\alpha) p_2) + 1}{n+1}\right)\\ + (1 - \alpha p_1 - (1-\alpha) p_2) V^N\left(n+1, \frac{n (\alpha p_1 + (1-\alpha) p_2)}{n+1}\right).
\end{multline}

The right-hand sides of \eqref{eqn:VbarConvex2} and \eqref{eqn:VbarConvex3} are both convex combinations of $V^N(n+1,\hp)$ with the same weights, which suggests using Lemma~\ref{lem:chords} (with $\alpha \gets \alpha p_1 + (1-\alpha) p_2$) to compare them. With the two terms in \eqref{eqn:VbarConvex3} playing the roles of $f(x + (1-\alpha) \delta)$ and $f(x - \alpha \delta)$ in Lemma~\ref{lem:chords}, we find 
\begin{align*}
x &= (\alpha p_1 + (1-\alpha) p_2) \frac{n (\alpha p_1 + (1-\alpha) p_2) + 1}{n+1} + (1 - \alpha p_1 - (1-\alpha) p_2) \frac{n (\alpha p_1 + (1-\alpha) p_2)}{n+1}\\
&= \frac{n (\alpha p_1 + (1-\alpha) p_2)}{n+1} + \frac{\alpha p_1 + (1-\alpha) p_2}{n+1}\\
&= \alpha p_1 + (1-\alpha) p_2,
\end{align*}
and a similar calculation with \eqref{eqn:VbarConvex2} yields the same value for $x$. Furthermore, comparing the arguments of the first terms in \eqref{eqn:VbarConvex2} and \eqref{eqn:VbarConvex3}, 
\begin{align*}
    &\frac{n}{n+1} \frac{\alpha p_1^2 + (1-\alpha) p_2^2}{\alpha p_1 + (1-\alpha) p_2} + \frac{1}{n+1} - \frac{n (\alpha p_1 + (1-\alpha) p_2) + 1}{n+1}\\ 
    &\qquad = \frac{n}{n+1} \frac{\alpha p_1^2 + (1-\alpha) p_2^2 - (\alpha p_1 + (1-\alpha) p_2)^2}{\alpha p_1 + (1-\alpha) p_2}\\
    &\qquad \geq 0,
\end{align*}
where the inequality is due to the convexity of the function $p^2$. This indicates that the $\delta$ corresponding to \eqref{eqn:VbarConvex2} (which will not be computed explicitly) is greater than or equal to the $\delta$ corresponding to \eqref{eqn:VbarConvex3}. Lemma~\ref{lem:chords} then implies that the right-hand side of \eqref{eqn:VbarConvex2} is greater than or equal to the right-hand side of \eqref{eqn:VbarConvex3}, thus completing the proof of \eqref{eqn:VbarConvex1}. (Note that this proof of convexity only required $V^N(n+1,\hp)$ to be convex in $\hp$, not necessarily non-decreasing.)

\subsection{Proof of Proposition 3}%\ref{prop:V*n}}
\label{sec:proofs:V*n}

First the base case is proven, i.e.~$V^N(N,\hp) \geq V^N(N+1, \hp)$, where $V^N(N+1,\hp) = V^*(\infty,\hp) = \max\{\hp-c, 0\} / (1-\gamma)$ and $V^N(N,\hp)$ is given by recursion \eqref{eqn:V*2} (with $V^*$ replaced by $V^N$). There are three cases corresponding to where the arguments on the right-hand side of \eqref{eqn:V*2}, $(N\hp)/(N+1)$ and $(N\hp+1)/(N+1)$, fall with respect to the threshold $c$. 

Case $(N\hp+1)/(N+1) \leq c$: Since this implies $(N\hp)/(N+1) < c$ and $\hp \leq c$, we have 
\[
V^N\left(N+1, \frac{N\hp + 1}{N+1} \right) = V^N\left(N+1, \frac{N\hp}{N+1} \right) = 0
\]
and the right-hand side of \eqref{eqn:V*2} yields $V^N(N,\hp) = 0$. This is equal to $V^N(N+1,\hp) = 0$.

Case $(N\hp)/(N+1) > c$: This implies $(N\hp+1)/(N+1) > c$ and $\hp > c$ as well. $V^N(N+1,\hp)$ is then a linear function over the interval $[(N\hp)/(N+1), (N\hp+1)/(N+1)]$ and 
\[
\hp V^N\left(N+1, \frac{N\hp + 1}{N+1} \right) + (1-\hp) V^N\left(N+1, \frac{N\hp}{N+1} \right) = V^N(N+1, \hp) = \frac{\hp - c}{1-\gamma}.
\]
Eq.~\eqref{eqn:V*2} then gives 
\[
V^N(N,\hp) = \max\left\{ \hp - c + \gamma \times \frac{\hp - c}{1-\gamma}, 0 \right\} = \frac{\hp - c}{1-\gamma} = V^N(N+1,\hp).
\]

Case $(N\hp)/(N+1) \leq c < (N\hp+1)/(N+1)$: Only one of the $V^N(N+1,\cdot)$ terms in \eqref{eqn:V*2} is non-zero, resulting in 
\begin{align*}
    V^N(N,\hp) &= \max\left\{ \hp - c + \frac{\gamma \hp}{1-\gamma} \left(\frac{N\hp + 1}{N+1} - c\right), 0 \right\}\\ 
    &= \max\left\{ \underbrace{(\hp - c) \left(1 + \frac{\gamma \hp}{1-\gamma}\right) + \frac{\gamma \hp (1-\hp)}{(1-\gamma)(N+1)} }_{\tV(N,\hp)}, 0 \right\}.
\end{align*}
In comparison,
\[
V^N(N+1,\hp) = \max\left\{ \tV(N+1,\hp), 0 \right\}, \quad \tV(N+1,\hp) = \frac{\hp - c}{1-\gamma}.
\]
Subtracting, 
\begin{align*}
    \tV(N,\hp) - \tV(N+1,\hp) &= (\hp - c) \frac{\gamma (\hp-1)}{1-\gamma} + \frac{\gamma \hp (1-\hp)}{(1-\gamma)(N+1)}\\
    &= \frac{\gamma (1-\hp)}{1-\gamma} \left(c - \hp + \frac{\hp}{N+1} \right)\\
    &\geq 0
\end{align*}
because $(N\hp)/(N+1) \leq c$ for this case. It follows that $V^N(N,\hp) \geq V^N(N+1,\hp)$.

Now for the inductive step, assume that $V^N(n,\hp) \geq V^N(n+1,\hp)$. Then
\begin{align*}
    V^N(n-1,\hp) &= \max\left\{ \hp - c + \gamma \left[\hp V^N\left(n, \hp + \frac{1 - \hp}{n}\right) + (1-\hp) V^N\left(n, \hp - \frac{\hp}{n}\right) \right], 0 \right\}\\
    &\geq \max\left\{ \hp - c + \gamma \left[\hp V^N\left(n+1, \hp + \frac{1 - \hp}{n}\right) + (1-\hp) V^N\left(n+1, \hp - \frac{\hp}{n}\right) \right], 0 \right\}\\
    &\geq \max\left\{ \hp - c + \gamma \left[\hp V^N\left(n+1, \hp + \frac{1 - \hp}{n+1}\right) + (1-\hp) V^N\left(n+1, \hp - \frac{\hp}{n+1}\right) \right], 0 \right\}\\
    &= V^N(n,\hp),
\end{align*}
where the second inequality follows from the convexity of $V^N(n+1,\hp)$ in $\hp$ (Proposition~\ref{prop:V*p}) and application of Lemma~\ref{lem:chords} with $x = \hp$, $\alpha = \hp$, and $\delta_1 = 1/n > \delta_2 = 1/(n+1)$.

\subsection{Proof of Theorem~\ref{thm:avg}}
\label{sec:proofs:avg}

Recall that in each belief state $(n,\hp)$, there are two choices for the acceptance probability $\pi = \Pi(n,\hp)$: either stop ($\pi = 0$) with zero reward $V^\Pi(n,\hp) = 0$, or accept with some positive probability, in which case $V^\Pi(n,\hp)$ is given by \eqref{eqn:VpiAvg}. To determine the optimal action by dynamic programming, we assume that an optimal policy is used from state $n+1$ onward, thus replacing $V^\Pi(n+1,\cdot)$ by $V^*(n+1,\cdot)$ on the right-hand side of \eqref{eqn:VpiAvg}. It follows that $\pi > 0$ is optimal if this right-hand side is non-negative. This in turn is true if $V^*(n+1,\hp)$ is convex in $\hp$ and non-negative, since Jensen's inequality would imply 
\begin{equation}\label{eqn:V*nAvg}
    V^\Pi(n,\hp) = \hp V^*\left(n+1, \frac{n\hp + 1}{n+1}\right) + (1-\hp) V^*\left(n+1, \frac{n\hp}{n+1}\right)
    \geq V^*(n+1,\hp)
    \geq 0.
\end{equation}

It is now shown by induction over decreasing $n$ that $V^*(n,\hp)$ is convex in $\hp$ and non-negative for all $n$, implying by the previous argument that $\Pi(n,\hp) > 0$ is optimal for all states. More precisely, we again consider approximations $V^N(n,\hp)$ to $V^*(n,\hp)$, initialized by setting $V^N(N+1,\hp) = V^*(\infty,\hp) = \max\{\hp-c, 0\}$. By taking $N \to \infty$, the proof extends to optimal policies.

The base case $n = N+1$ is simply given by the initialization $V^N(N+1,\hp) = \max\{\hp - c, 0\}$, since this is a convex and non-negative function. It then suffices to establish convexity for $n = N, N-1, \dots$ since \eqref{eqn:V*nAvg} would then show that $V^N(n,\hp)$ is non-increasing in $n$, not just non-negative. This inductive step corresponds exactly with the proof of convexity of the function $\bV(n,\hp)$ in the proof of Proposition~\ref{prop:V*p} (Appendix~\ref{sec:proofs:V*p}).

\fi

\end{document}